\documentclass{article}

\usepackage[preprint]{corl_2026} 

\usepackage{graphics,graphicx,float,subcaption,booktabs,xcolor,multirow,array,color,ifthen,colortbl,dblfloatfix,url,xparse,mathtools,patchcmd,amssymb,xspace,nicefrac,microtype,amsmath,amsthm,amsfonts,bm,ragged2e,tikz,stackengine,etoolbox,xpatch,enumerate,xstring,setspace,tabularx,makecell,changepage,titlesec,enumitem,wrapfig,tcolorbox,gensymb,algorithm,algpseudocode,makecell,bbm}
\usepackage{titlesec}
\titlespacing*{\section}{0pt}{0.6\baselineskip}{0.5\baselineskip}
\definecolor{CMURed}{HTML}{A6192E}
\hypersetup{bookmarksopen,bookmarksnumbered,
pdfpagemode=UseOutlines,
colorlinks=true,
linkcolor=blue,
anchorcolor=blue,
citecolor=blue,
filecolor=blue,
menucolor=blue,
urlcolor=CMURed
}
\usepackage[capitalise,noabbrev,nameinlink]{cleveref}
\usepackage[english]{babel}
\usepackage[font=small]{caption}

\newcommand{\link}[1]{\url{#1}}

\newcommand{\website}{https://linchangyi1.github.io/ART-Glove}
\newcommand{\ProjectWeb}[0]{\href{\website}{\website}}

\usepackage{xspace}
\newcommand{\nickname}[0]{ART-Glove\xspace}

\title{\nickname: Articulated Tactile Glove for \\Contact-Grounded Dexterous Interaction Capture}
\author{
Changyi Lin, Ding Zhao\\
[0.15cm]
Carnegie Mellon University\\
[0.15cm]
\ProjectWeb
}

\begin{document}
\maketitle

\begin{figure}[h]
    \centering
    \includegraphics[width= \linewidth]{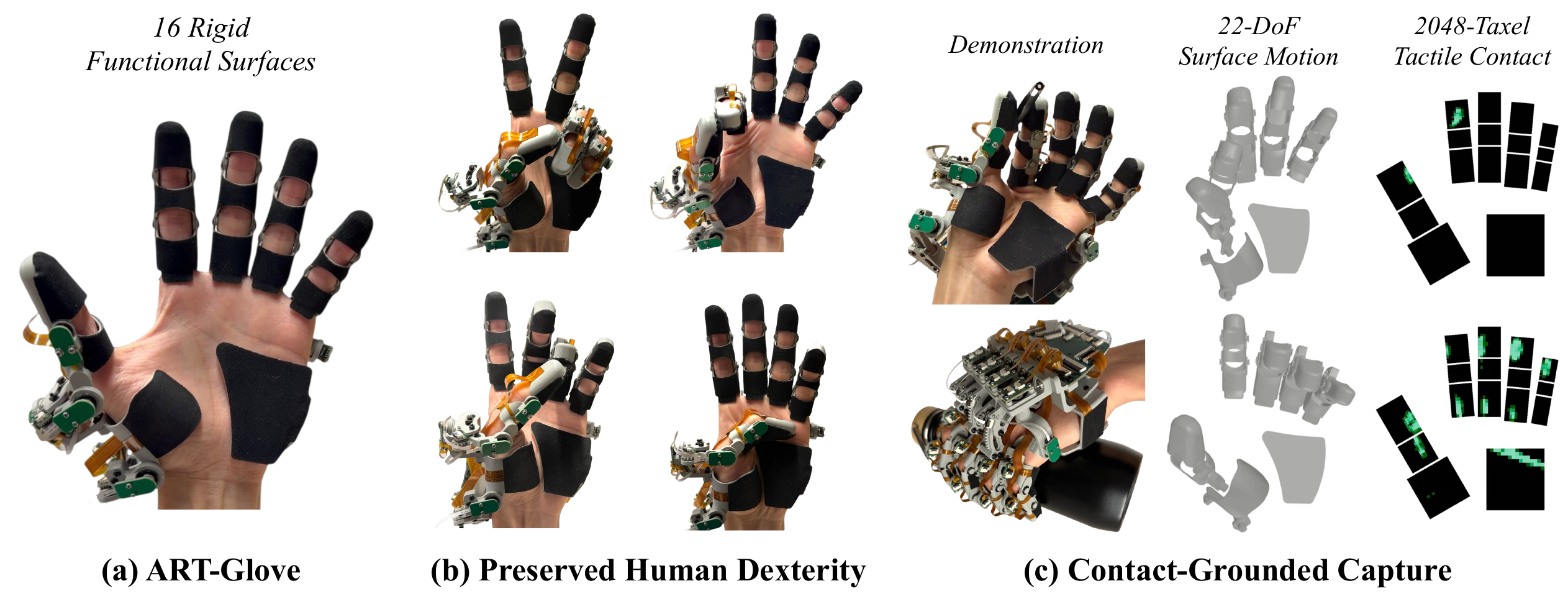}
    \caption{\small \nickname enables contact-grounded dexterous demonstration capture. (a) \nickname represents the hand with 16 rigid functional surfaces, making the hand-side contact geometry explicit. (b) Anatomically aligned articulation preserves human dexterity across natural hand poses. (c) \nickname records synchronized contact-grounded information, including 22-DoF surface motion and dense 2048-taxel tactile contact signals.}
    \label{fig:teaser}
\end{figure}


\begin{abstract}
We present \nickname, an articulated tactile glove designed to capture contact-grounded dexterous demonstrations while preserving human dexterity. \nickname makes hand-side contact geometry explicit with 16 rigid functional surfaces covering the fingers, thumb, and palm. Twenty-two anatomically aligned joints connect these surfaces and allow them to follow human hand motion during dexterous manipulation. Encoder-based sensing tracks surface motion, while dense piezoresistive tactile sensing records contact over the same surfaces. The complete system captures synchronized 22-DoF joint measurements and 2048-taxel tactile measurements at 120~Hz. We evaluate \nickname across experiments on motion freedom, joint sensing, tactile sensing, and contact-rich interaction capture, demonstrating its ability to preserve human dexterity while recording contact-grounded information that can support downstream dexterous robot learning.
\end{abstract}


\section{Introduction}
\label{sec:intro}

Dexterous manipulation remains a central challenge in robotics due to its contact-rich interactions and high-dimensional hand control. Human demonstrations provide a natural source of supervision, as they capture expert full-hand coordination across diverse manipulation tasks. Learning from such demonstrations involves two coupled stages: humans first perform tasks to generate interaction data, and the resulting data are then converted into supervision for downstream robot learning.

This pipeline imposes requirements on both sides of the capture process. On the acquisition side, the method should preserve \emph{human dexterity}: the demonstrator should retain natural hand motion, effective haptic feedback, and usable full-hand contact area. On the data side, the recorded demonstration should provide \emph{contact-grounded information}. For contact-rich manipulation, hand kinematics alone are insufficient to specify the underlying physical interaction: the same motion can produce different outcomes depending on which hand-side surface contacts the object, the local surface geometry, how that surface moves, and when and where contact occurs. A useful demonstration should therefore make this physical interaction observable by providing known or reliably recoverable hand-side contact geometry, tracked surface motion, and measured or inferred contact state.

Existing demonstration capture methods satisfy these requirements to different degrees, as summarized in Table~\ref{tab:dexterous_data_collection}. Teleoperation yields robot-ready data and policy action labels, but teleoperating contact-rich tasks remains difficult due to limited haptic feedback to the operator and embodiment mismatch. Rigid exoskeletons and linkage-based interfaces provide more explicit physical surfaces and mechanical force feedback, but their bulky structures, reduced degrees of freedom, or kinematic misalignment can constrain natural full-hand interaction. Bare-hand videos preserve natural dexterity and scale easily, but hand geometry, surface motion, and contact state must be inferred from visual observations; these inferences become unreliable under occlusion and contact. Soft gloves add sensing while preserving motion, but their deformable and user-dependent surfaces remain difficult to represent as accurate contact geometry. These methods therefore face a trade-off between preserving human dexterity and recording contact-grounded information.

We present \nickname, an articulated tactile glove designed to occupy a complementary point in this trade-off. The key idea is to make the human hand-side interaction surface explicit while preserving natural dexterous behavior. \nickname represents the hand with 16 rigid phalanx- and palm-level functional surfaces, providing known contact geometry over the fingers, thumb, and palm. These surfaces are connected by 22 anatomically aligned joints so that they follow human hand motion while preserving near-free dexterity. Encoder-based joint sensing tracks surface motion, and dense piezoresistive tactile sensing records contact over the same surfaces. The complete system captures synchronized 22-DoF joint measurements and 2048-taxel tactile measurements at 120~Hz.

With this design, \nickname enables humans to perform dexterous demonstrations naturally while recording contact-grounded information, including surface geometry, surface motion, and tactile contact signals. We evaluate \nickname as a hardware platform for contact-grounded dexterous interaction capture, focusing on preserved human dexterity, joint-angle sensing accuracy, integrated tactile response, and full-system synchronized interaction capture.

\begin{table}[t]
\centering
\setlength{\tabcolsep}{2pt}
\caption{Comparison of dexterous data collection methods.}
\vspace{0.1cm}
\resizebox{\linewidth}{!}{
\begin{tabular}{lccccccc}
\toprule
& \multicolumn{3}{c}{\textbf{Human Dexterity}}
& \multicolumn{3}{c}{\textbf{Contact-Grounded Information}}
& \multirow{3}{*}{\makecell{\textbf{Action}\\\textbf{Label}}}
\\
\cmidrule(lr){2-4}
\cmidrule(lr){5-7}
& \begin{tabular}[c]{c} \textbf{Motion}\\ \textbf{Freedom} \end{tabular}
& \begin{tabular}[c]{c} \textbf{Haptic}\\ \textbf{Feedback} \end{tabular}
& \begin{tabular}[c]{c} \textbf{Functional}\\ \textbf{Area} \end{tabular}
& \begin{tabular}[c]{c} \textbf{Contact}\\ \textbf{Geometry} \end{tabular}
& \begin{tabular}[c]{c} \textbf{Surface}\\ \textbf{Motion} \end{tabular}
& \begin{tabular}[c]{c} \textbf{Tactile}\\ \textbf{Sensing} \end{tabular}
\\
\midrule
\makecell{
Teleoperation\\~\cite{cheng2024open,ding2025bunny,zhang2025doglove,du2025mile}}
& \makecell{Robot-\\constrained}
& \makecell{Indirect/\\limited}
& Robot hand
& \makecell{Known\\robot geometry}
& \makecell{Robot\\state}
& \makecell{Robot-\\dependent}
& \makecell{Robot\\commands}
\\
\midrule
\makecell{Fingertip\\Exoskeleton~\cite{xu2025dexumi,si2025exostart}}
& Constrained
& \makecell{Surface-\\mediated}
& Fingertips
& \makecell{Known\\device geometry}
& \makecell{Encoder\\ (12-DoF)}
& \makecell{Sparse /\\optional}
& \makecell{Next\\motion}
\\
\midrule
\makecell{Passive-Hand\\Linkage~\cite{fang2025dexop,zhu2026whed,zhu2026dexexo}}
& Constrained
& \makecell{Linkage-\\mediated}
& \makecell{Low-DoF\\ passive hand}
& \makecell{Known\\robot geometry}
& \makecell{Encoder\\ (12-DoF)}
& \makecell{Varies /\\optional}
& \makecell{Next\\motion}
\\
\midrule
\makecell{Bare-Hand Video\\~\cite{hoque2025egodex,kareer2025egomimic,zheng2026egoscale,punamiya2026egoverse}}
& Free
& Direct
& Full hand
& \makecell{Estimated\\deformable hand}
& \makecell{Vision\\ (21-DoF)}
& \makecell{None /\\inferred}
& \makecell{Next\\motion}
\\
\midrule
\makecell{Soft Sensing Gloves\\~\cite{wang2024dexcap,tao2025dexwild,song2025opentouch,yin2025osmo}}
& Near-free
& \makecell{Near-\\direct}
& Full hand
& \makecell{Estimated\\deformable glove}
& \makecell{Glove-\\dependent}
& \makecell{Sparse /\\optional}
& \makecell{Next\\motion}
\\
\midrule
\makecell{\textbf{\nickname{}}\\\textbf{(Ours)}}
& \textbf{Near-free}
& \makecell{\textbf{Surface-}\\\textbf{mediated}}
& \textbf{Full hand}
& \makecell{\textbf{Known}\\\textbf{rigid surfaces}}
& \makecell{\textbf{Encoder}\\\textbf{(22-DoF)}}
& \makecell{\textbf{Dense}\\\textbf{(2048 taxels)}}
& \makecell{\textbf{Next}\\\textbf{motion}}
\\
\bottomrule
\end{tabular}}
\par\vspace{0.2cm}
\begin{minipage}{0.98\linewidth}
\footnotesize{\textit{Note.} ``Surface-mediated'' means that the user contacts objects through instrumented glove or device surfaces while still receiving mechanical contact feedback. ``Linkage-mediated'' means that contact forces are transmitted through a linkage from the passive robot hand. ``Next motion'' denotes human/device motion labels.}
\end{minipage}
\label{tab:dexterous_data_collection}
\end{table}

\section{Related Work}
\label{sec:related_work}

\textbf{Teleoperation.}
Teleoperation~\cite{handa2020dexpilot,qin2023anyteleop,zhao2023learning,cheng2024open} is popular for robot data collection because it directly records robot-side observations and policy actions. However, contact-rich dexterous teleoperation remains difficult because visual observations are often occluded and haptic feedback is limited. Prior systems have explored fingertip force feedback~\cite{schwarz2021nimbro,zhang2025doglove,tang2026towards}, vibrotactile cues~\cite{ding2025bunny,gao2025glovity}, and localized pneumatic tactile displays~\cite{jia2026feel}. While these methods provide useful feedback channels, they are typically sparse, localized, or less intuitive for contact-rich full-hand manipulation. In contrast, \nickname allows humans to directly manipulate objects through full-hand surface-mediated contact.

\textbf{Wearable Interfaces.}
Handheld sensorized grippers have shown that robot-free data collection can be scaled in the wild~\cite{chi2024universal}. Extending this idea to dexterous hands is attractive, because wearable interfaces can let humans physically drive robot-like hand surfaces and receive more direct haptic feedback than remote teleoperation. Exoskeleton-based systems attach robot-consistent surface structures to the human hand and constrain human motion toward robot-feasible configurations~\cite{xu2025dexumi,si2025exostart}, while linkage-based systems allow human hands to drive passive robot-like hands through mechanical linkages~\cite{fang2025dexop,zhu2026whed,zhu2026dexexo}. However, these methods can constrain natural hand motion and limit full-hand functional interaction due to reduced DoFs, bulky mechanisms, or kinematic mismatch. In contrast, \nickname uses anatomically aligned articulation to preserve near-free human motion while maintaining full-hand functional surfaces.

\textbf{Human Videos and Soft Gloves.}
Bare-hand videos provide scalable human demonstrations with natural dexterity and full-hand interaction. Prior works learn dexterous priors from internet videos~\cite{shaw2023videodex,shaw2024learning}, use egocentric videos for policy pretraining or co-training with robot data~\cite{hoque2025egodex,yang2025egovla,kareer2025egomimic,niu2025human2locoman,zheng2026egoscale,punamiya2026egoverse,li2026egolive}, or reconstruct hand-object trajectories for policy learning~\cite{chen2025vividex,chen2026dexterous}. However, video-based methods must infer deformable hand geometry, hand motion, and contact states from visual observations, which becomes unreliable under contact and occlusion. Soft wearable gloves add hand-pose or tactile sensing while largely preserving natural motion~\cite{wang2024dexcap,tao2025dexwild,song2025opentouch,yin2025osmo}, but their soft and user-dependent surfaces remain difficult to represent accurately. In contrast, \nickname uses rigid articulated surfaces to provide known contact geometry while preserving natural human motion.

\textbf{Tactile Sensing.}
Vision-based tactile sensors provide rich local contact information~\cite{yuan2017gelsight,lin20239dtact,ward2018tactip,zhang2022deltact}, but their camera-based structures are too thick to preserve haptic feedback in wearable full-hand interfaces. Piezoresistive sensors are thinner and more scalable, making them better suited for dense wearable tactile coverage~\cite{sundaram2019learning}. Recent systems have improved piezoresistive tactile sensing through FPCB-based sensor fabrication~\cite{murphy2025fits,huang2026flexitac} and low-crosstalk readout electronics~\cite{johnson2025scaling,lin2026hipi}. \nickname builds on this line of work by integrating dense piezoresistive tactile sensing into an articulated full-hand glove to capture contact signals together with known surface geometry and surface motion.


\begin{figure}
\centering
\includegraphics[width=\linewidth]{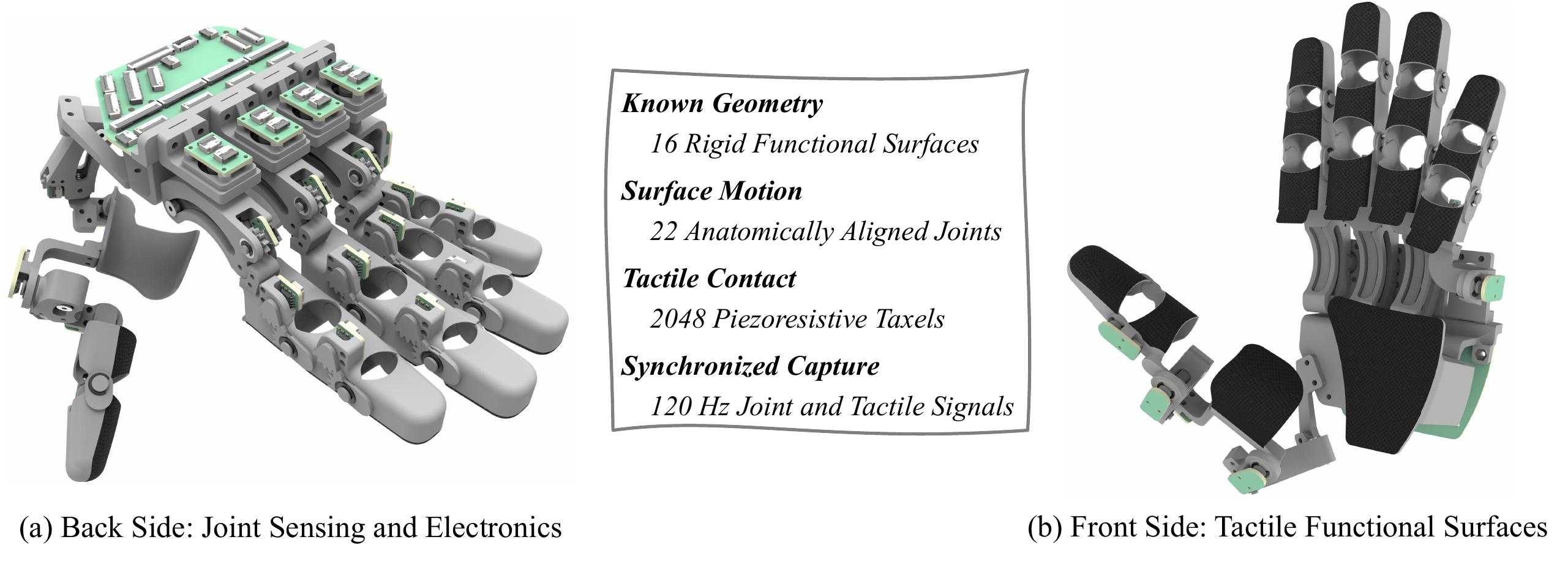}
\caption{\small System overview of \nickname. 
(a) The back side integrates articulated joint mechanisms, encoder-based joint sensing, and electronics. 
(b) The front side exposes tactile-covered rigid functional surfaces for full-hand interaction. The system contains 16 functional rigid surfaces, 22 anatomically aligned joints, 2048 piezoresistive taxels, and synchronized signal acquisition at 120 Hz.}
\label{fig:overview}
\end{figure}

\section{Design Requirements and System Overview}
\label{sec:system_overview}

\nickname is designed to capture contact-grounded information from dexterous human demonstrations while preserving human dexterity. This goal leads to four system-level requirements. First, the glove should provide known hand-side contact geometry through rigid functional surfaces. Second, these surfaces should follow the corresponding human hand segments without substantially restricting natural motion. Third, contact should be measured over the same functional surfaces. Fourth, surface-motion and tactile-contact signals should be synchronized in a wearable capture system.

As shown in Fig.~\ref{fig:overview}, \nickname consists of an articulated rigid-shell glove and a compact sensing stack. The front side exposes tactile-covered rigid functional surfaces over the fingers, thumb, and palm. The back side integrates the joint mechanisms, encoder PCBs, sensor hub, readout PCB, and MCU board. The following mechanical design section (Sec.~\ref{sec:mechanical_design}) describes how the rigid surfaces and anatomical articulation are realized, while the sensing design section (Sec.~\ref{sec:sensing_design}) describes how tactile contact, joint motion, and synchronized acquisition are implemented.

\section{Mechanical Design of \nickname}
\label{sec:mechanical_design}

\subsection{Rigid Functional Surfaces}

\begin{wrapfigure}{r}{0.27\textwidth}
\vspace{-0.5cm}
\centering
\includegraphics[width=\linewidth]{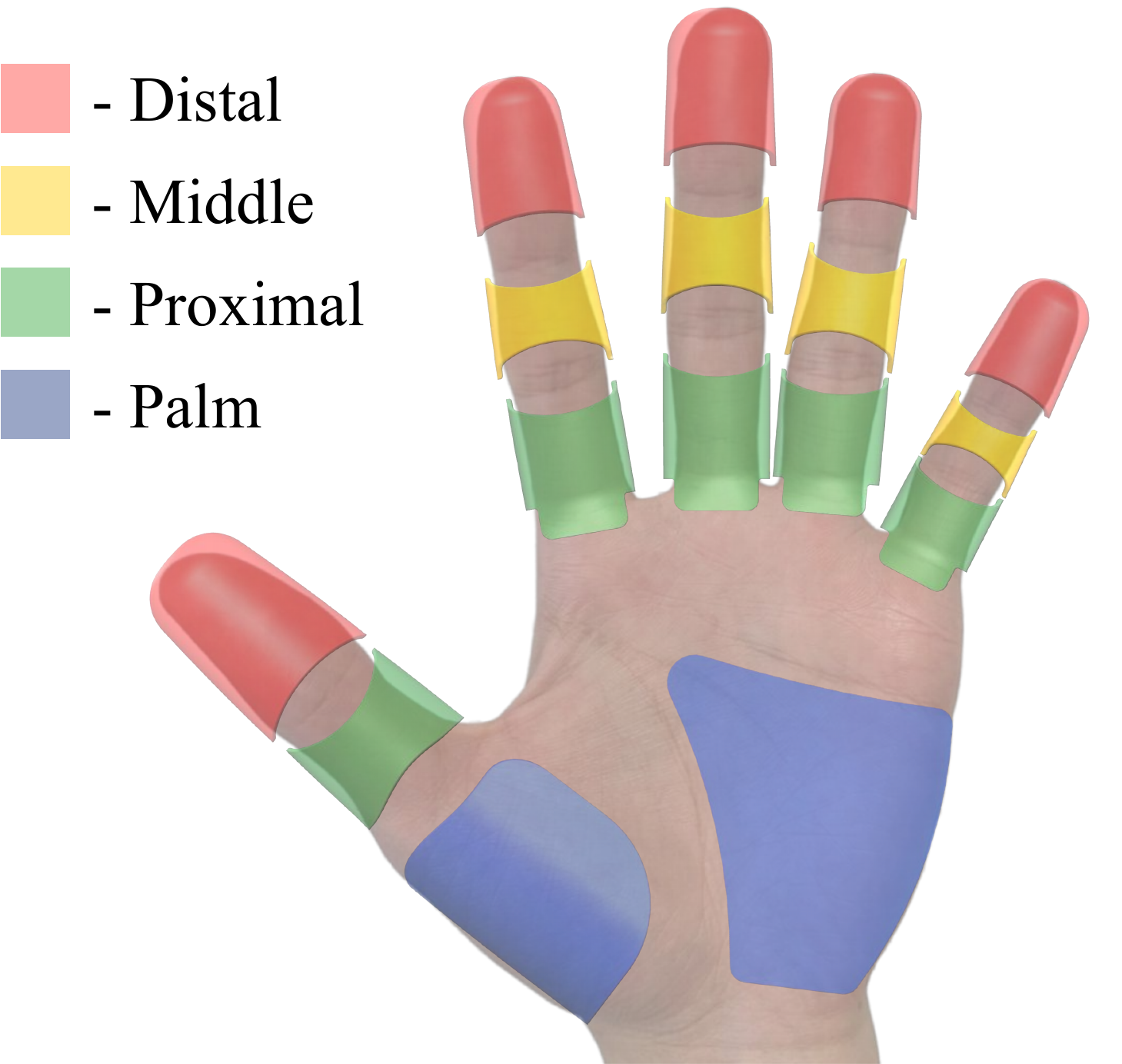}
\vspace{-0.5cm}
\caption{\small Rigid functional surfaces defined by \nickname.}
\label{fig:functional_surfaces}
\end{wrapfigure}

As illustrated in Fig.~\ref{fig:functional_surfaces}, \nickname represents the human hand using a set of rigid functional surface links, which define the hand-side contact geometry used during manipulation. Each of the four fingers is divided into three phalanx-level surfaces corresponding to the distal, middle, and proximal phalanges. The thumb is represented by three surfaces corresponding to the distal phalanx, proximal phalanx, and first metacarpal. The remaining palm contact region is covered by a rigid palm shell. Together, these 16 rigid functional surfaces over the fingers, thumb, and palm provide known contact geometry for full-hand dexterous interaction capture. Regions between adjacent surfaces are left open to provide clearance for joint motion and finger bending.

\subsection{Anatomically Aligned Joints}

\subsubsection{From Fingertip Tracking to Full-Surface Capture}

For fingertip-oriented exoskeletons~\cite{zhang2025doglove,xu2025dexumi,zhu2026whed,jia2026feel}, exact alignment between device and human joint axes is not always necessary, because the fingertip pose can be reconstructed from an external kinematic chain with offset joints. In contrast, \nickname targets full-surface interaction capture, where each rigid phalanx link serves as a functional contact surface and should move with the corresponding human hand segment. Using offset external joints would require additional linkages to recover the motion of each rigid surface link, substantially increasing mechanical complexity and constraining natural hand motion. Therefore, \nickname adopts anatomically aligned joints, allowing each rigid surface link to directly follow the corresponding human hand segment.

\subsubsection{Anatomical Joint Arrangement}

\begin{wrapfigure}{r}{0.27\textwidth}
\vspace{-1.6cm}
\centering
\includegraphics[width=\linewidth]{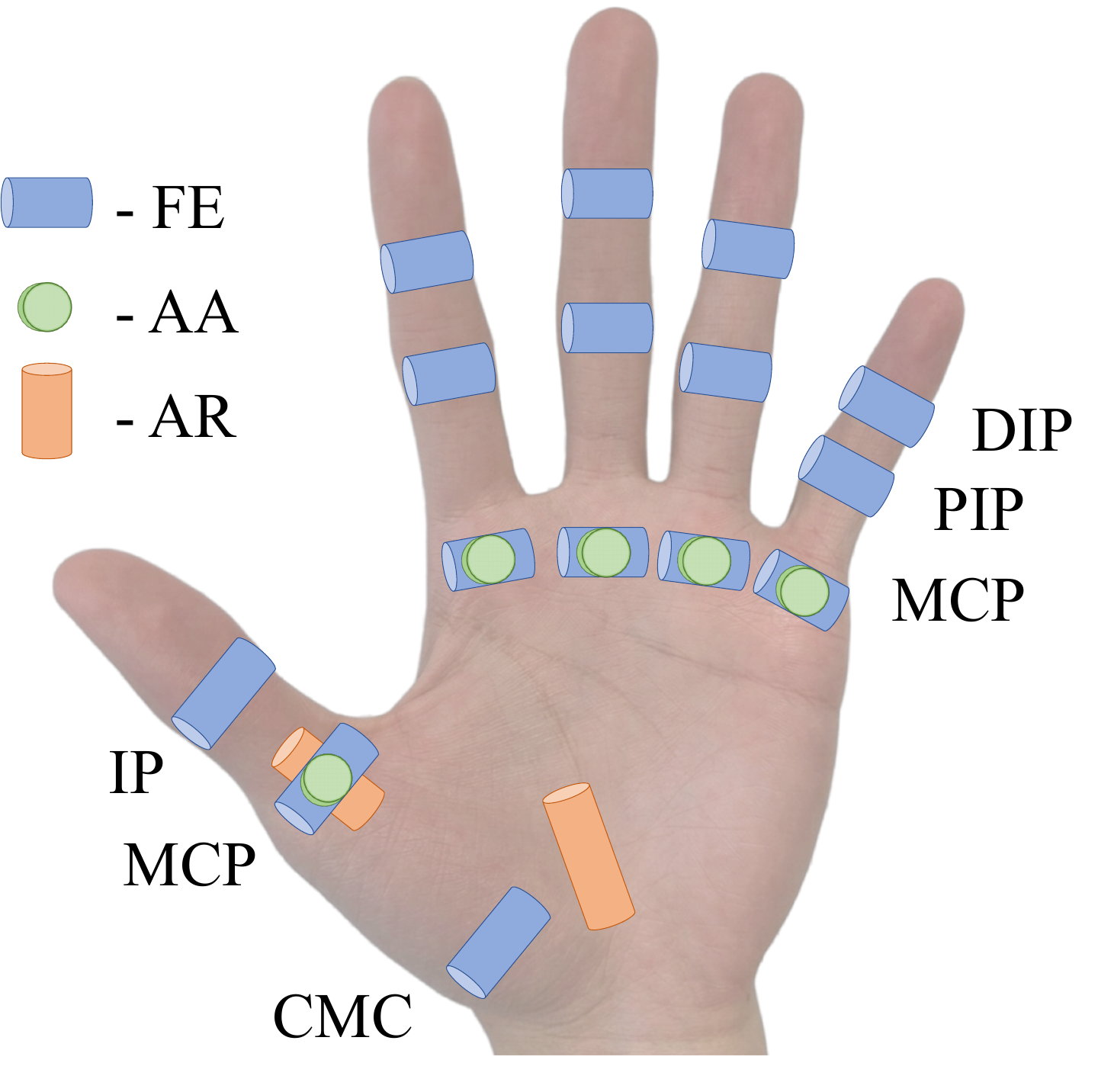}
\vspace{-0.5cm}
\caption{\small Anatomical arrangement of the 22 joints.}
\vspace{-0.5cm}
\label{fig:joint_arrangement}
\end{wrapfigure}

As illustrated in Fig.~\ref{fig:joint_arrangement}, \nickname uses a 22-DoF anatomical joint arrangement to describe the motion of its rigid surface links. Each of the four fingers is modeled with four DoFs: DIP flexion/extension, PIP flexion/extension, MCP flexion/extension (MCP-FE), and MCP abduction/adduction (MCP-AA). The thumb is modeled with six DoFs: IP-FE, MCP-FE, MCP-AA, MCP axial rotation (MCP-AR), CMC-FE, and CMC axial rotation (CMC-AR). In total, the four fingers contribute $4 \times 4 = 16$ DoFs, and the thumb contributes 6 DoFs.

While many robotic hands simplify thumb motion by placing axial rotation only near the CMC region, \nickname includes axial-rotation DoFs at both the MCP and CMC regions to better track the motion of thumb contact surfaces. MCP-AR captures the coupled pronation/supination component associated with thumb MCP motion, while CMC-AR captures base thumb rotation associated with folding and opposition. This design is motivated by biomechanical studies~\cite{hollister1995axes,li2007coordination}, which show that thumb motion arises from coordinated multi-axis rotations across the CMC, MCP, and IP joints.

\subsubsection{Spatially Adaptive Joint Mechanisms}

\begin{figure}
\centering
\includegraphics[width=\linewidth]{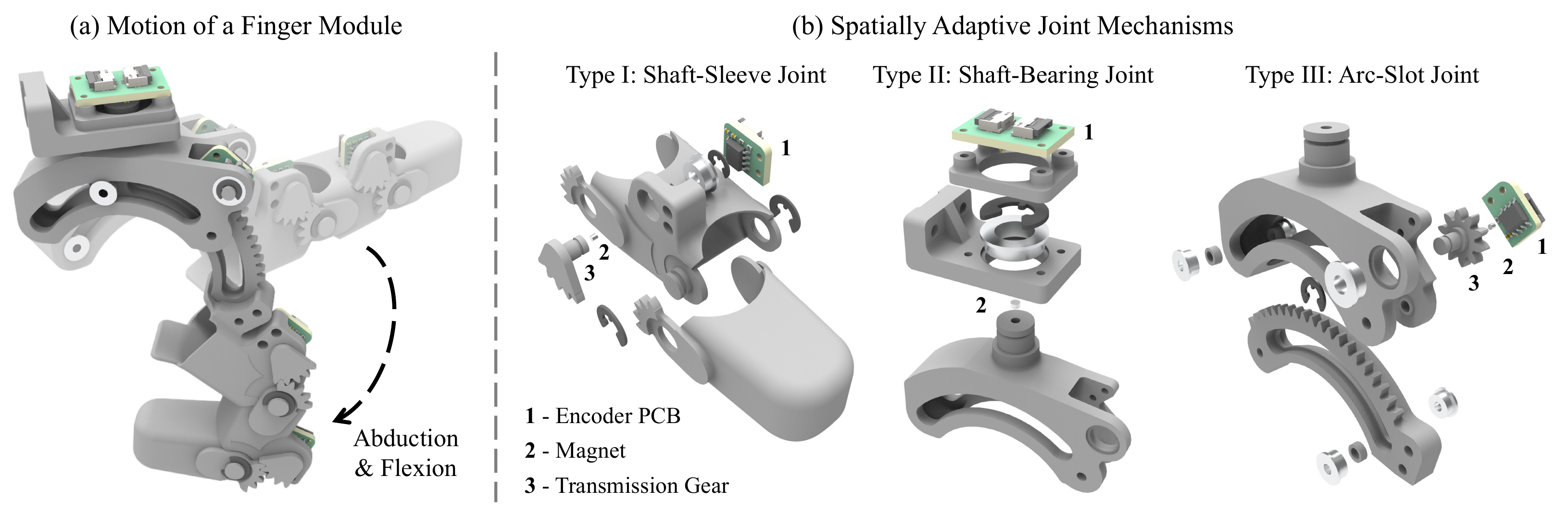}
\caption{\small Joint mechanism design of \nickname. (a) An index-finger module illustrates how the articulated shell follows finger motion. (b) Three spatially adaptive mechanisms realize anatomically aligned axes under different local clearance constraints: Type I shaft-sleeve joints, Type II shaft-bearing joints, and Type III arc-slot joints.}
\label{fig:joint_mechanisms}
\end{figure}

Realizing anatomically aligned joints requires placing each mechanism near the desired anatomical axis while preserving clearance for hand motion and hand-object contact. The available lateral clearance varies across the hand. DIP, PIP, and thumb IP joints are relatively isolated and provide usable clearance on both sides of the joint axis. MCP-AA joints and several thumb joints have asymmetric access, where only one side can accommodate glove components. Other axes provide little lateral clearance on either side, such as the MCP-FE joints of the central fingers near the palm. These local constraints motivate three spatially adaptive joint mechanisms, as illustrated in Fig.~\ref{fig:joint_mechanisms}.

\textbf{Type I: Shaft-sleeve joints for two-side-clearance axes.}
DIP and PIP axes provide access on both sides, but the joint profile must remain compact to preserve inter-finger clearance for MCP-AA motion. We therefore avoid separate bearings and integrate the shaft and sleeve directly into adjacent rigid links, with metal E-clips constraining axial motion. This design reduces part count and limits the lateral protrusion to $3.7\,\mathrm{mm}$ from the inner glove surface to the outer joint profile on each side. For finger DIP and PIP joints, rotation is transmitted through gears to a compact sensing axis on the back of the finger, allowing the encoder PCB and magnet to be mounted without increasing inter-finger width. The same shaft-sleeve mechanism is also used for the thumb IP joint, where the encoder can be mounted directly on the less-constrained outer side.

\textbf{Type II: Shaft-bearing joints for one-side-clearance axes.}
Several MCP and thumb joints provide sufficient clearance on one accessible side of the anatomical axis. For these axes, we use a shaft-bearing joint to improve smoothness and support. A shaft is aligned with the anatomical axis and supported by a miniature bearing on the accessible side, where the encoder PCB and magnet are also integrated for angle measurement. We use shaft-bearing joints for the MCP-AA joints of all five digits, the MCP-FE joints of the pinky and thumb, and the thumb CMC-FE joint.

\textbf{Type III: Arc-slot joints for no-side-clearance axes.}
When neither side of the anatomical axis provides enough lateral clearance for a side-mounted shaft, we use an arc-slot joint. This occurs near the palm, where adjacent fingers are tightly packed, and at thumb axial-rotation joints, where a side-mounted support would interfere with surrounding structures. The arc-slot joint realizes the desired rotation axis without placing a shaft directly on that axis: small bearings roll inside curved slots whose center is aligned with the anatomical axis. To improve stability, each support point uses a symmetric pair of bearings, and two separated arc slots increase the support baseline. Because the anatomical axis is not directly accessible for sensing, link motion is transmitted to a smaller gear whose axis can accommodate the encoder PCB and magnet. We use arc-slot joints for the MCP-FE joints of the three central fingers and for the thumb MCP-AR and CMC-AR joints.

\section{Sensing and Data Acquisition}
\label{sec:sensing_design}

\begin{figure}
\centering
\includegraphics[width=\linewidth]{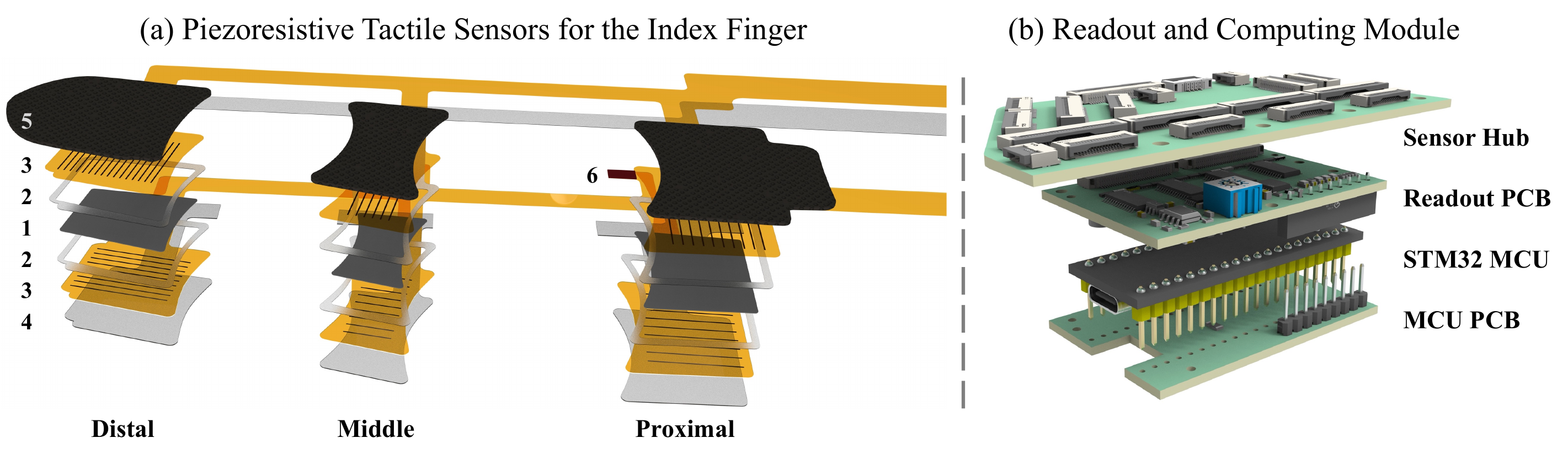}
\caption{\small Sensing and readout design of \nickname.
(a) Piezoresistive tactile modules for the index finger cover the distal, middle, and proximal rigid surface links. Numbers indicate key layers and interfaces: 1 -- piezoresistive layer, 2 -- 3M9077 tape, 3 -- conductive layer, 4 -- 3M468 tape, 5 -- cover layer, and 6 -- FPCB connector. (b) Readout and computing module. Distributed tactile and encoder signals are routed through the sensor hub and readout PCB to the STM32-based acquisition board.}
\label{fig:sensing_design}
\end{figure}
\vspace{-0.2cm}

\subsection{Full-Hand Tactile Sensing}

\nickname uses piezoresistive tactile sensing to measure contact over its rigid functional surfaces. The 16 rigid functional surfaces are covered by a full-hand tactile skin with $64 \times 32 = 2048$ sensing units, or \textit{taxels}. To simplify routing and assembly, the tactile skin is organized into seven modular FPCBs covering the fingers, thumb, and palm. 
Fig.~\ref{fig:sensing_design}(a) shows the tactile modules for the index finger, which cover the distal, middle, and proximal rigid surface links. Each tactile module has a layered structure consisting of a piezoresistive layer, conductive FPCB electrodes, adhesive layers, and a protective cover layer. A taxel is formed at each row--column electrode intersection. When contact is applied, the local resistance of the piezoresistive layer changes and is measured through the corresponding electrode pair. The cover layer protects the electrodes and provides a high-friction contact surface. The 3M9077 tape separates and bonds the conductive layers, while the 3M468 tape attaches the tactile module to the rigid shell.

\subsection{Joint Angle Sensing}

\nickname measures joint angles using compact contactless magnetic rotary encoders. Each sensing unit consists of an AS5600L magnetic rotary position sensor and a diametrically magnetized disc magnet. The magnet is mechanically coupled to the joint rotation, and the encoder PCB measures the absolute angular position from the magnetic field. As shown in Fig.~\ref{fig:joint_mechanisms}, the magnet is placed either directly on the anatomical joint axis or on a mechanically coupled sensing axis, depending on the local joint mechanism. Encoder signals are routed through the FPCB connectors of the corresponding tactile modules, reducing additional wiring around the articulated links.

\subsection{Signal Synchronization and Readout}

The tactile modules connect to the sensor hub shown in Fig.~\ref{fig:sensing_design}(b), which aggregates distributed tactile and encoder signals from the glove. The readout PCB scans the structured tactile array using a low-crosstalk readout design~\cite{lin2026hipi}, while the STM32F411CE MCU collects tactile measurements through SPI and encoder measurements through I2C. This pipeline produces synchronized joint and tactile signals at $120\,\mathrm{Hz}$. The PCBs are compactly stacked and mounted on the back of the glove. More details on the tactile layout, FPCB design, and readout circuit are provided in the Appendix.

\section{Experiments}
\label{sec:Experiments}

We evaluate \nickname along three dimensions that characterize its ability to capture contact-grounded dexterous demonstrations. First, we assess preserved human dexterity through mechanical range-of-motion measurements and representative finger and thumb motions. Second, we assess sensing capability through joint-angle sensing accuracy and tactile response. Third, we demonstrate integrated interaction capture during contact-rich manipulation tasks, where \nickname records synchronized surface motion and tactile contact signals.

\subsection{Motion Freedom}

\vspace{-0.4cm}
\begin{table}[h]
\centering
\footnotesize
\setlength{\tabcolsep}{5pt}
\caption{Mechanical range of motion of \nickname.}
\vspace{0.1cm}
\begin{tabular}{lcccccc}
\toprule
\textbf{Finger Joint} 
& DIP 
& PIP 
& MCP-FE 
& MCP-AA 
& \multicolumn{2}{c}{}
\\
\textbf{ROM} 
& $[0^{\circ}, 70^{\circ}]$
& $[0^{\circ}, 75^{\circ}]$
& $[0^{\circ}, 88^{\circ}]$
& $[-90^{\circ}, 90^{\circ}]$
& \multicolumn{2}{c}{}
\\
\midrule
\textbf{Thumb Joint} 
& IP
& MCP-FE 
& MCP-AA 
& MCP-AR 
& CMC-FE 
& CMC-AR 
\\
\textbf{ROM} 
& $[-65^{\circ}, 70^{\circ}]$
& $[-30^{\circ}, 90^{\circ}]$
& $[-30^{\circ}, 30^{\circ}]$
& $[0^{\circ}, 30^{\circ}]$
& $[-30^{\circ}, 90^{\circ}]$
& $[0^{\circ}, 25^{\circ}]$
\\
\bottomrule
\end{tabular}
\label{tab:mechanical_rom}
\end{table}

\begin{figure}
\centering
\includegraphics[width=\linewidth]{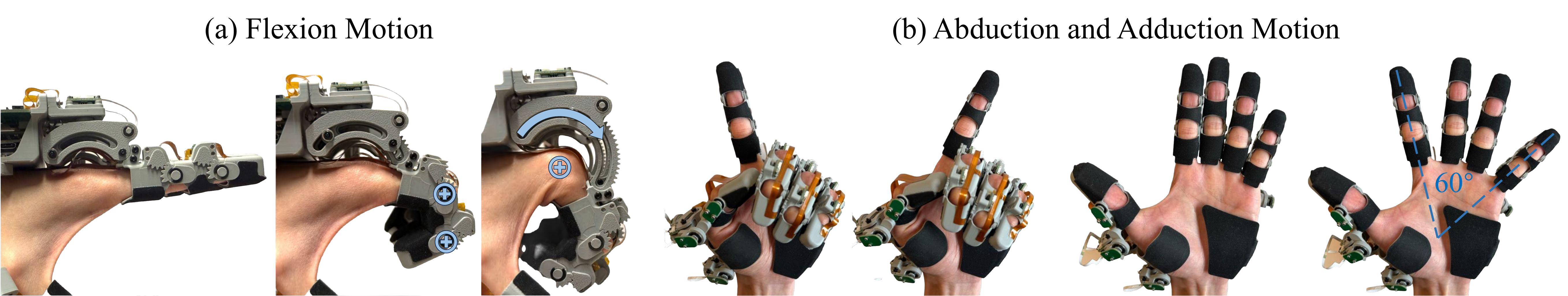}
\caption{\small Finger motion enabled by \nickname.
(a) Flexion sequence showing the finger moving from extension to flexion while the articulated shell follows the finger surface.
(b) Abduction/adduction sequence showing lateral finger motion between adjacent fingers while maintaining full-hand tactile functional surfaces.}
\label{fig:finger_motion}
\end{figure}

We first evaluate whether \nickname provides sufficient motion freedom for dexterous demonstrations. Table~\ref{tab:mechanical_rom} reports the mechanical range of motion (ROM) of each joint type. For the four fingers, \nickname provides flexion ranges of $70^{\circ}$ at DIP, $75^{\circ}$ at PIP, and $88^{\circ}$ at MCP-FE. These ranges allow the finger module to move continuously from an extended posture to a deeply flexed posture while the rigid surface links follow the corresponding phalanx regions, as shown in Fig.~\ref{fig:finger_motion}(a).

For finger abduction/adduction, the MCP-AA joint has a nominal mechanical range of $[-90^{\circ}, 90^{\circ}]$. In practice, the effective spreading motion is mainly limited by neighboring finger modules rather than internal joint stops. Fig.~\ref{fig:finger_motion}(b) shows that \nickname preserves a large range of finger abduction and adduction: the spread angle between the index finger and pinky exceeds $60^{\circ}$.

\begin{figure}
\centering
\vspace{-0.1cm}
\includegraphics[width=\linewidth]{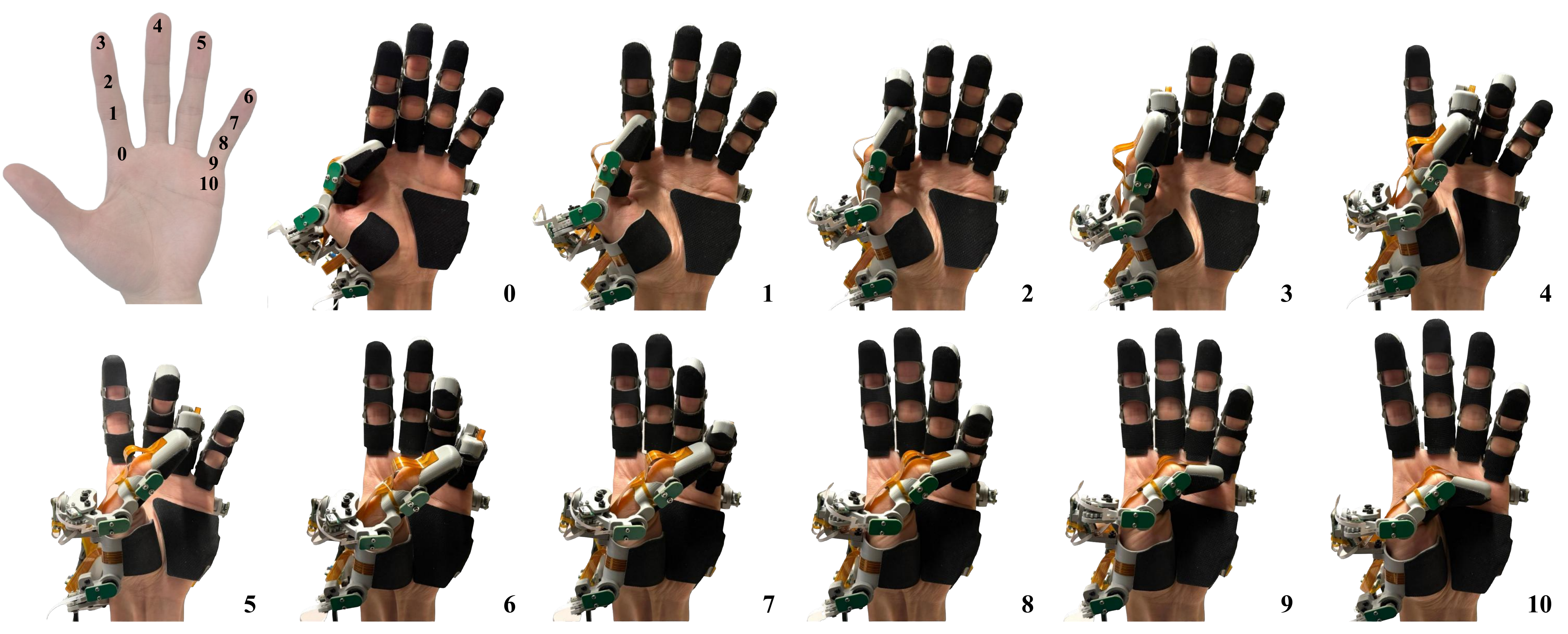}
\caption{\small Thumb motion evaluated with the Kapandji opposition test. The thumb reaches all positions from 0 to 10, indicating that the articulated thumb module preserves a wide range of opposition motion.}
\label{fig:thumb_k_test}
\vspace{-0.2cm}
\end{figure}

For the thumb, \nickname provides an IP range of $[-65^{\circ}, 70^{\circ}]$ and additional MCP and CMC DoFs for flexion/extension, lateral positioning, axial rotation, and opposition-related folding. We further evaluate thumb mobility using the Kapandji opposition test, which measures the ability of the thumb to reach different regions of the hand. As shown in Fig.~\ref{fig:thumb_k_test}, the wearer reaches all positions while wearing \nickname, achieving a Kapandji score of 10, which indicates that the six-DoF thumb module preserves a wide range of thumb opposition for dexterous interaction.

\subsection{Sensing Capability}
\subsubsection{Joint Sensing Accuracy}

\begin{figure}[t]
\centering
\includegraphics[width=\linewidth]{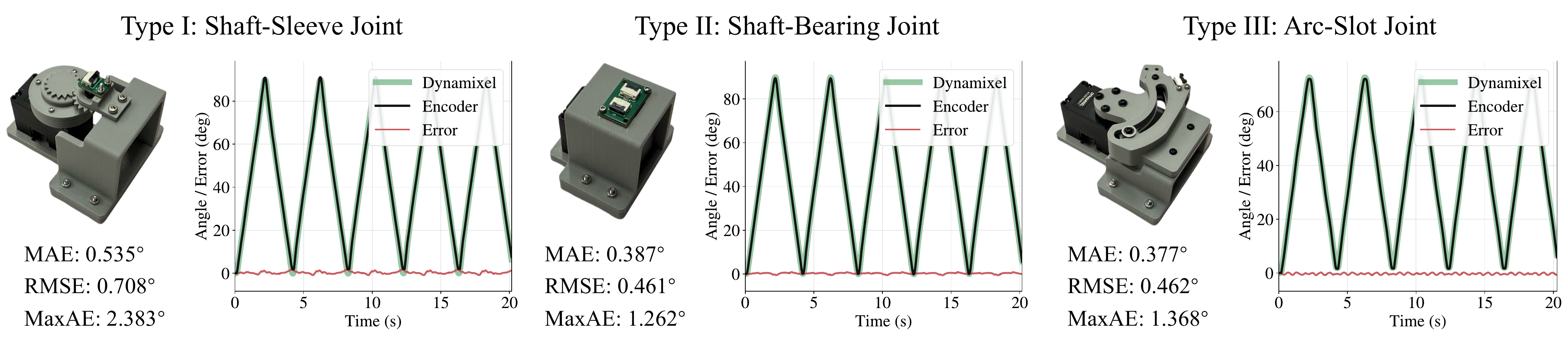}
\caption{\small
Joint-angle sensing accuracy of the three joint mechanisms.
Each mechanism is mounted to a Dynamixel XC330 servo, which provides the
reference joint angle. All three mechanisms achieve sub-degree mean absolute error, with maximum absolute error below $2.4^\circ$.
}
\label{fig:joint_evaluation}
\vspace{-0.2cm}
\end{figure}

We evaluate the joint-angle sensing accuracy of the three joint mechanisms using the setup shown in Fig.~\ref{fig:joint_evaluation}. Following prior servo-based evaluations~\cite{du2025mile,jia2026feel}, each mechanism is mounted to a Dynamixel XC330 servo. The servo drives the joint through repeated bidirectional angle sweeps and provides the reference angle measurement. For each mechanism, we record five continuous forward--backward cycles and compare the \nickname encoder measurement with the servo reference. The Type I shaft-sleeve and Type II shaft-bearing joints are tested over a $0^\circ$--$90^\circ$ range, while the Type III arc-slot joint is tested over a $0^\circ$--$75^\circ$ range due to its smaller mechanical range of motion.

As shown in Fig.~\ref{fig:joint_evaluation}, all three mechanisms achieve sub-degree mean absolute error. The Type I shaft-sleeve joint shows the largest error, mainly due to the gear transmission between the joint axis and the sensing axis.  The Type II shaft-bearing joint achieves lower error because its sensing axis is directly coupled to the joint without gear transmission. The Type III arc-slot joint also uses gear transmission, but its $7.4{:}1$ transmission ratio increases the effective sensing resolution, resulting in comparable accuracy to the directly coupled Type II joint. These results indicate that the proposed joint mechanisms provide reliable joint-angle measurements.

\subsubsection{Integrated Tactile Sensing}

\begin{figure}[t]
\centering
\includegraphics[width=\linewidth]{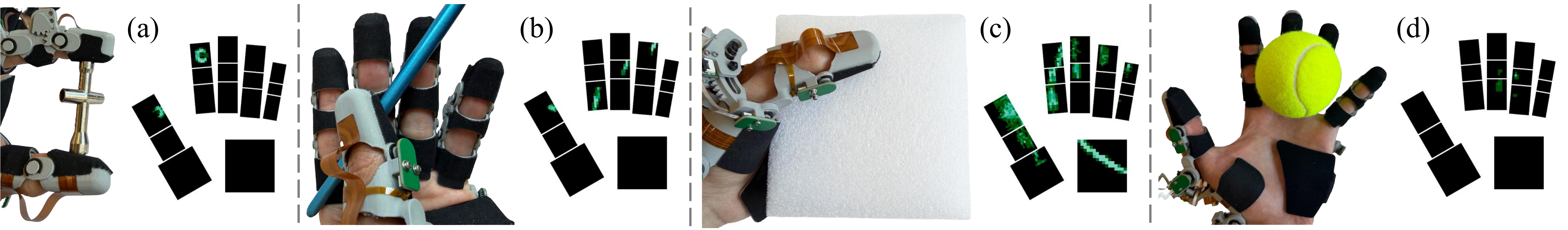}
\caption{\small Tactile sensing response of \nickname.
(a) A circular contact produces a localized shape response. (b) Pen contact activates only the contacting functional surfaces. (c) Grasping a soft foam block produces distributed full-hand contact. (d) A tennis ball resting under gravity produces clear light-contact activation.
}
\label{fig:tactile_demo}
\vspace{-0.2cm}
\end{figure}

We evaluate the tactile response of \nickname. As shown in Fig.~\ref{fig:tactile_demo}, the examples cover four properties of tactile sensing: contact shape, surface selectivity, distributed full-hand contact, and light-contact sensitivity. The wrench contact produces a localized circular activation pattern, showing that the tactile skin preserves the spatial structure of contact on the surface. The pen contact activates only the surfaces that are physically touched, while non-contacting surfaces remain inactive. This indicates that tactile activation is localized to the corresponding functional surfaces. When grasping a soft foam block, \nickname records distributed contact over multiple fingers and the palm, showing that the tactile skin captures full-hand contact patterns. Finally, a tennis ball resting on the hand produces clear activation without intentional pressing, indicating sensitivity to light contact. These examples show that the integrated tactile skin provides spatially localized and sensitive contact measurements over the rigid functional surfaces of \nickname.

\subsection{Dexterous Interaction Capture}
\label{sec:dexterous_interaction_capture}

\begin{figure}[t]
\centering
\includegraphics[width=\linewidth]{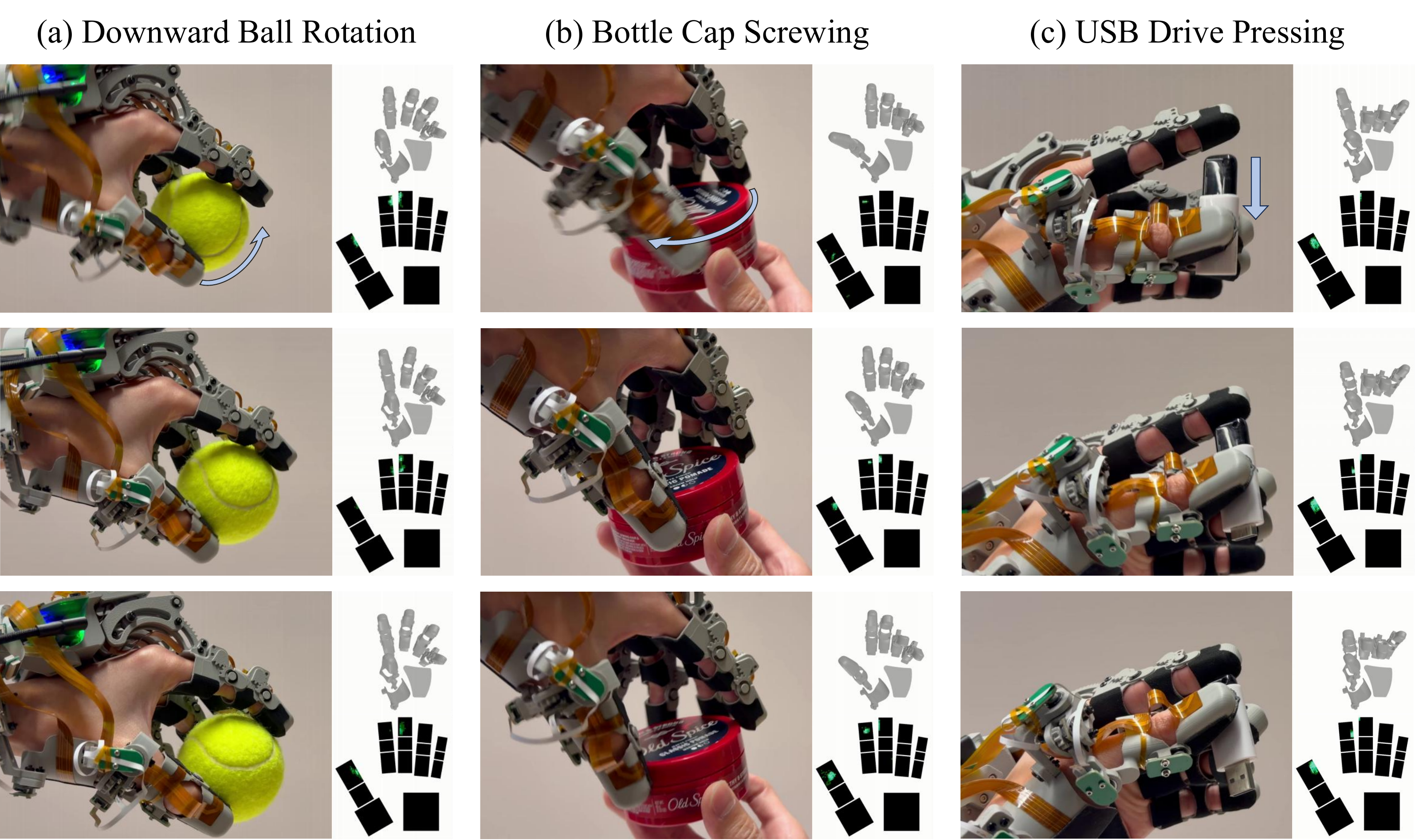}
\caption{\small Dexterous interaction capture with \nickname. We record contact-rich demonstrations across three representative tasks: (a) downward ball rotation, (b) bottle-cap screwing, and (c) USB-drive pressing. Each example shows the demonstration together with 22-DoF surface motion and 2048-taxel tactile contact.}
\label{fig:demo}
\vspace{-0.4cm}
\end{figure}

We demonstrate full-system interaction capture on three contact-rich tasks, as shown in Fig.~\ref{fig:demo}. The tasks are selected to stress both dexterous execution and contact-grounded recording. In downward ball rotation, the wearer must reorient the ball while maintaining contact against gravity, which requires continuous contact regulation and benefits from both motion freedom and haptic feedback. In bottle-cap screwing, the wearer rapidly rotates the cap while maintaining stable contact. The tactile maps show changing contact locations and increased activation near the end of the screwing motion. In USB-drive pressing, the thumb and middle finger stabilize the drive while the index finger presses it, demonstrating coordinated multi-finger interaction with simultaneous grasping and localized actuation. Across these demonstrations, \nickname records the motion of its rigid surfaces through 22-DoF joint measurements and measures tactile contact over the same surfaces. The synchronized surface-motion and tactile-contact streams provide contact-grounded demonstrations at 120~Hz.

\section{Conclusion}
\label{sec:conclusion}

We presented \nickname, an articulated tactile glove for contact-grounded dexterous interaction capture. \nickname represents the human hand with 16 rigid functional surfaces, connects these surfaces through 22 anatomically aligned joints, and integrates encoder-based joint sensing with dense piezoresistive tactile sensing. This design makes hand-side contact geometry explicit while allowing the wearer to perform dexterous manipulation with preserved human dexterity. Experiments show that \nickname provides sufficient motion freedom for finger and thumb motion, achieves reliable joint-angle sensing across its three joint mechanisms, captures tactile responses over the integrated functional surfaces, and records synchronized surface-motion and tactile-contact signals during contact-rich demonstrations. These results support \nickname as a hardware platform for collecting contact-grounded human demonstrations that can support dexterous robot learning.

\section{Limitations and Discussion}

\textbf{Rigid Shell.}
The current rigid shells are 1-mm-thick PLA parts fabricated by 3D printing. Although this thickness keeps the glove lightweight, thinner shells could provide additional clearance for hand motion and further improve dexterity. Besides, shell stiffness affects tactile robustness: local shell deformation can introduce tactile artifacts, which we observe on the thumb first-metacarpal surface where soft-tissue motion and shell deformation are relatively large. Stronger materials could enable thinner shells while reducing deformation-induced tactile noise. In addition, the current fingertip shells do not include nail-like structures, making small-object pickup more difficult. Adding rigid nail extensions could improve interactions with small objects.

\textbf{Joint Mechanisms.}
The Type III arc-slot mechanism realizes rotation about an anatomical axis without placing a shaft or support on either side of the joint. However, the current implementation relies on low-cost bearings, screw--nut fasteners, and 3D-printed slots and gears, which make the mechanism relatively bulky. Its performance is also sensitive to assembly tolerance: over-tightening increases friction, while under-tightening causes lateral wobble of the guided components and reduces motion accuracy. More compact and standardized arc-guided mechanisms, such as curved guided rails, could improve the accuracy, wearability, and robustness of \nickname.

\textbf{Tactile Sensing.}
The tactile skin of \nickname has several limitations. First, the current piezoresistive skin only measures distributed normal contact magnitudes. Force calibration across curved surfaces remains future work. Second, the current tactile coverage is limited to the front functional surfaces of the glove. In practice, side surfaces of the fingers and thumb can also be heavily used during dexterous manipulation. Vision-based tactile sensors could provide richer contact information, but existing designs are still too bulky for dense full-hand wearable integration. Future thinner, multi-axis, and side-covering tactile sensors could further improve the tactile capability of \nickname.

\textbf{Skill Transfer from Gloves to Robots.}
\nickname is not intended for one-to-one skill transfer from a single glove to a specific robot hand. Human hands vary substantially in size and geometry, and a universal articulated glove that fits all users while preserving dexterity is difficult to realize. Instead, \nickname targets a \textit{many-to-many} transfer setting: demonstrations can be collected with customized gloves for different users, while robot hands can learn from a shared contact-grounded description of interaction. This description represents demonstrations through hand-side surface properties, including known geometry and nominal material, together with surface motion and contact state, rather than only joint trajectories. A robot hand therefore does not need to exactly match any individual glove; it should instead provide comparable information about its own contact surfaces, their motion, and contact state. One possible embodiment is an actuated hand based on a canonical \nickname-like geometry, such as an average geometry derived from the demonstration gloves. More broadly, as contact-grounded interaction data accumulate across devices and robots, learning can move from one-to-one joint matching toward modeling physical interaction between moving surfaces.

\textbf{Scalable Design and Fabrication.} 
The current prototype is designed for a specific hand size. Scaling \nickname to many demonstrators requires a design pipeline that can personalize both the mechanical structure and the sensing layout. This is more than geometric resizing: the rigid surface links should match each user's phalanx and palm geometry, the joint axes should remain aligned with the user's hand motion, and the tactile FPCBs should preserve consistent taxel-to-surface mappings across customized gloves. A promising direction is to generate these designs from hand scans captured by accessible devices such as smartphones or portable 3D scanners. The pipeline could then automatically produce fabrication files for the rigid shells, joint components, tactile FPCBs, and calibration tools. Such scalable fabrication would reduce manual design effort and enable larger multi-user datasets for contact-grounded dexterous interaction capture.

\section*{Acknowledgments}

The authors would like to thank Yuxiang Yang, Maria Bauza, Marissa Giustina, and Peide Huang from Google DeepMind for their valuable advice and discussions.

\newpage
\bibliography{main} 

\newpage
\appendix
\section{Additional Joint Mechanism Details}

\begin{figure}[h]
\centering
\includegraphics[width=\linewidth]{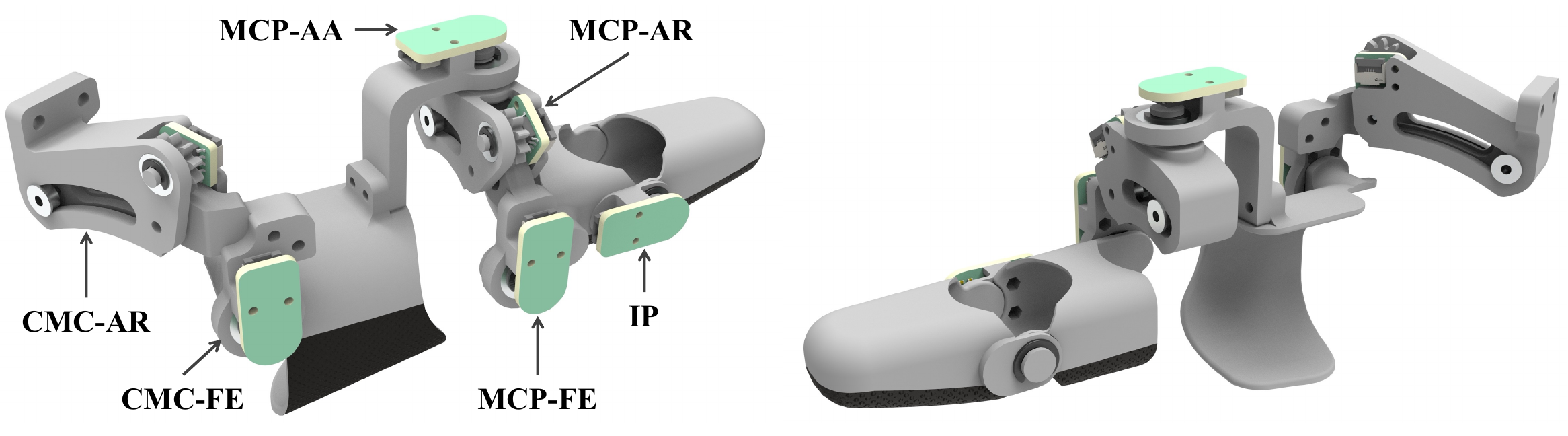}
\caption{\small Thumb module design of \nickname.}
\label{fig:thumb_design}
\end{figure}

The thumb has six DoFs, as shown in Fig.~\ref{fig:thumb_design}. The thumb MCP-FE joint adopts the Type II shaft-bearing mechanism because its inner side is connected to the palm, leaving usable space mainly on the outer side. The MCP-AA and MCP-AR joints are strongly coupled during thumb opposition, which can be observed in the high-score poses of the Kapandji test. Since axial rotation is unique to the thumb, the thumb fingertip must be firmly attached to the rigid shell so that human thumb rotation can drive the corresponding glove surface.

The CMC-AR and CMC-FE joints are important for enabling the thumb to reach positions 6--10 in the Kapandji test. The thumb metacarpal surface is not designed as a closed ring because the human first metacarpal is connected to the palm. As a result, CMC flexion can drive the rigid metacarpal surface by pushing against it, while CMC extension cannot reliably pull the surface back. In practice, we attach the thumb metacarpal region to the rigid surface with double-sided tape so that CMC extension can also drive the glove surface.

\vspace{-0.1cm}

\begin{figure}[h]
\centering
\includegraphics[width=0.7\linewidth]{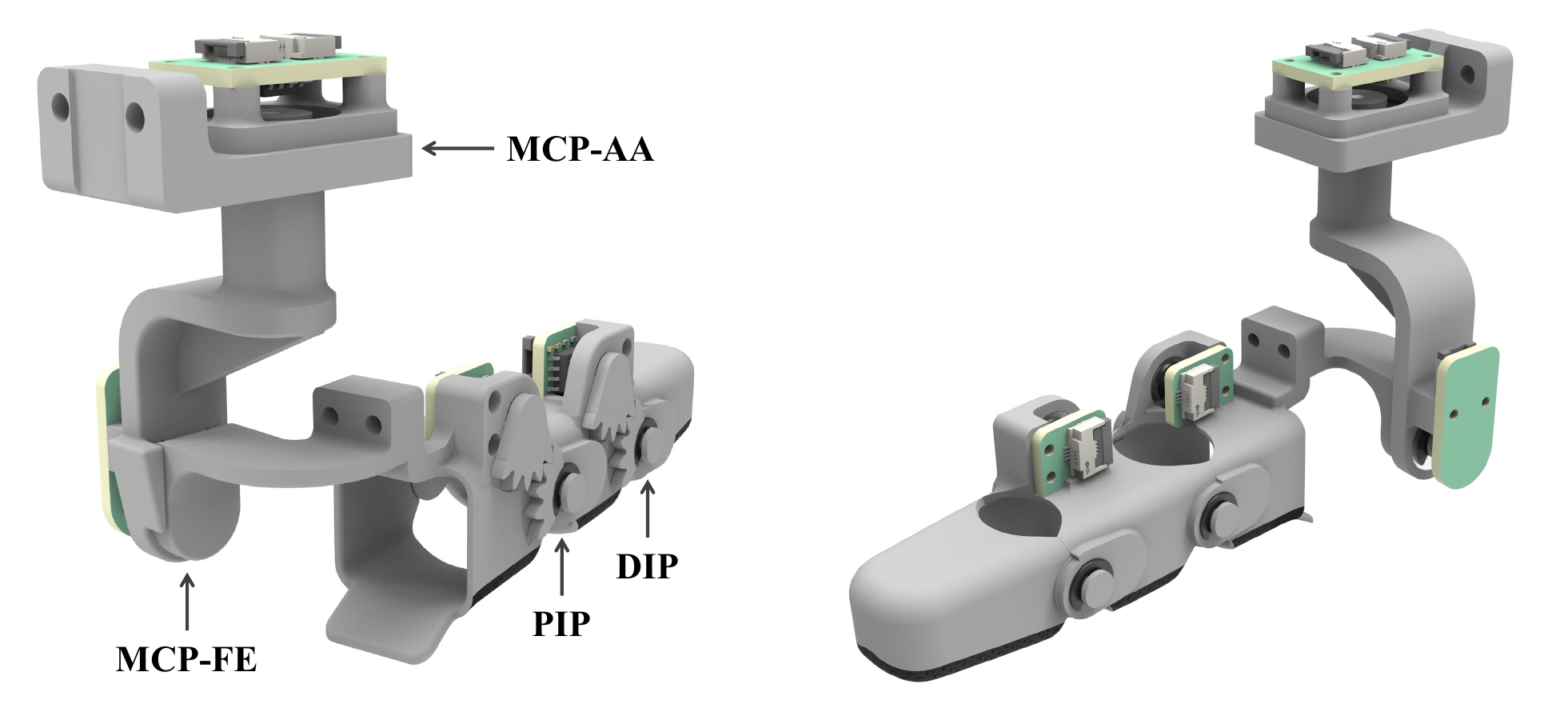}
\caption{\small Pinky module design of \nickname.}
\label{fig:pinky}
\end{figure}

The main difference between the pinky module and the other finger modules is the MCP-FE mechanism, as shown in Fig.~\ref{fig:pinky}. The pinky MCP-FE joint uses the Type II shaft-bearing mechanism because the outer side of the pinky provides usable space for a side-mounted joint. In addition, the spacing between the pinky and ring-finger bases is limited, making it difficult to place two Type III arc-slot mechanisms next to each other.

\vspace{-0.1cm}

\begin{figure}[t] \centering \includegraphics[width=0.7\linewidth]{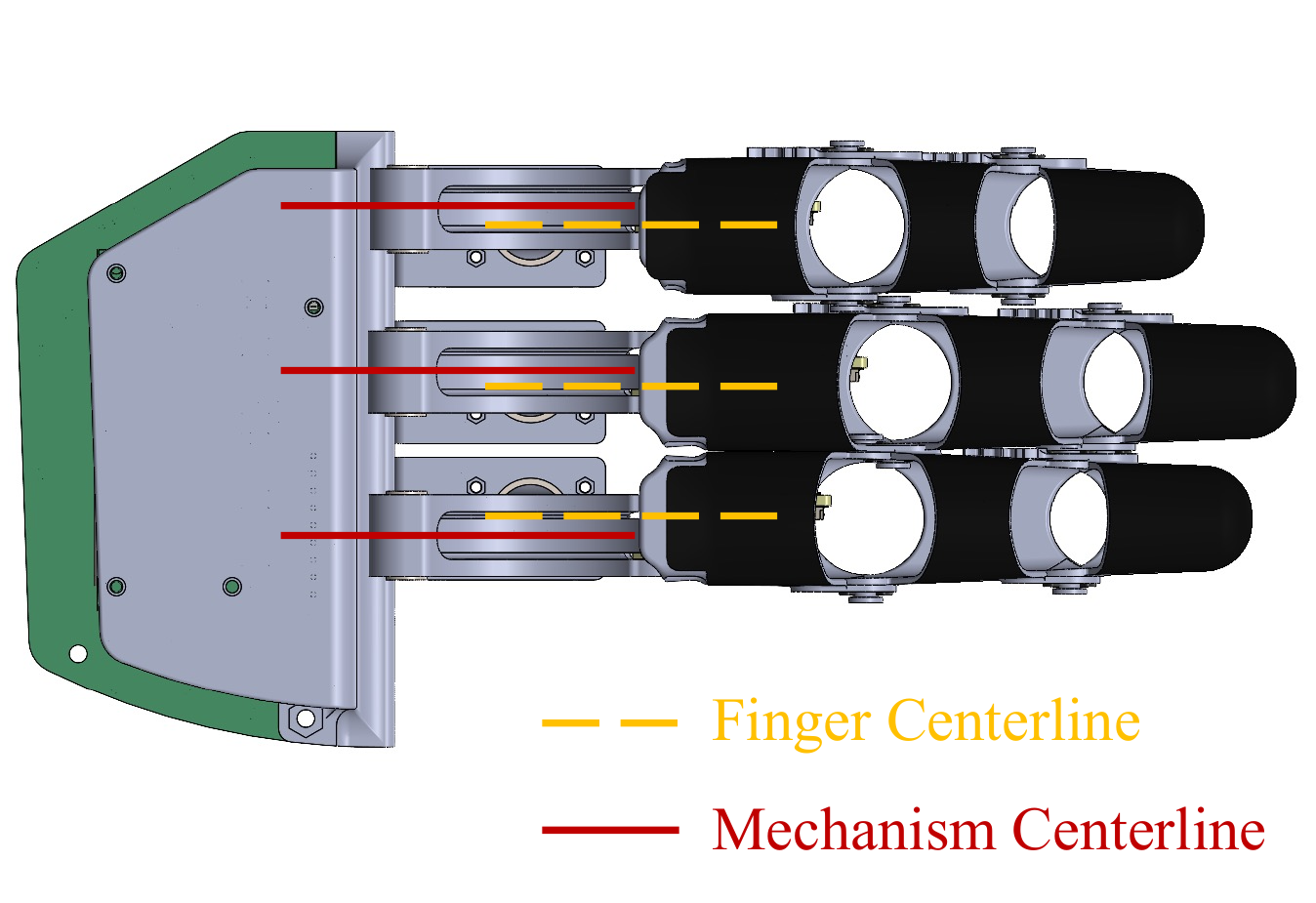}
\caption{\small Lateral offset of the Type III MCP-FE mechanism centerlines. Yellow dashed lines indicate finger centerlines, and red solid lines indicate mechanism centerlines. The offset balances the spacing between adjacent arc-slot mechanisms and preserves clearance for finger abduction/adduction. }
\label{fig:offset_joint_arrangement}
\end{figure}

As shown in Fig.~\ref{fig:offset_joint_arrangement}, the spacing between adjacent fingers is not uniform: the gap between the index and middle fingers is larger than the gap between the middle and ring fingers. If each Type III arc-slot mechanism were centered on its corresponding finger, the middle and ring mechanisms would be too close to each other and would restrict finger abduction/adduction. To preserve lateral finger motion, we laterally offset the mechanism centerlines so that the spacing between adjacent mechanism modules is more balanced. This offset improves clearance between the Type III mechanisms while keeping the rigid surface links aligned with the corresponding fingers.

\clearpage

\section{Taxel Layout}
\begin{figure}[h]
\centering
\includegraphics[width=0.6\linewidth]{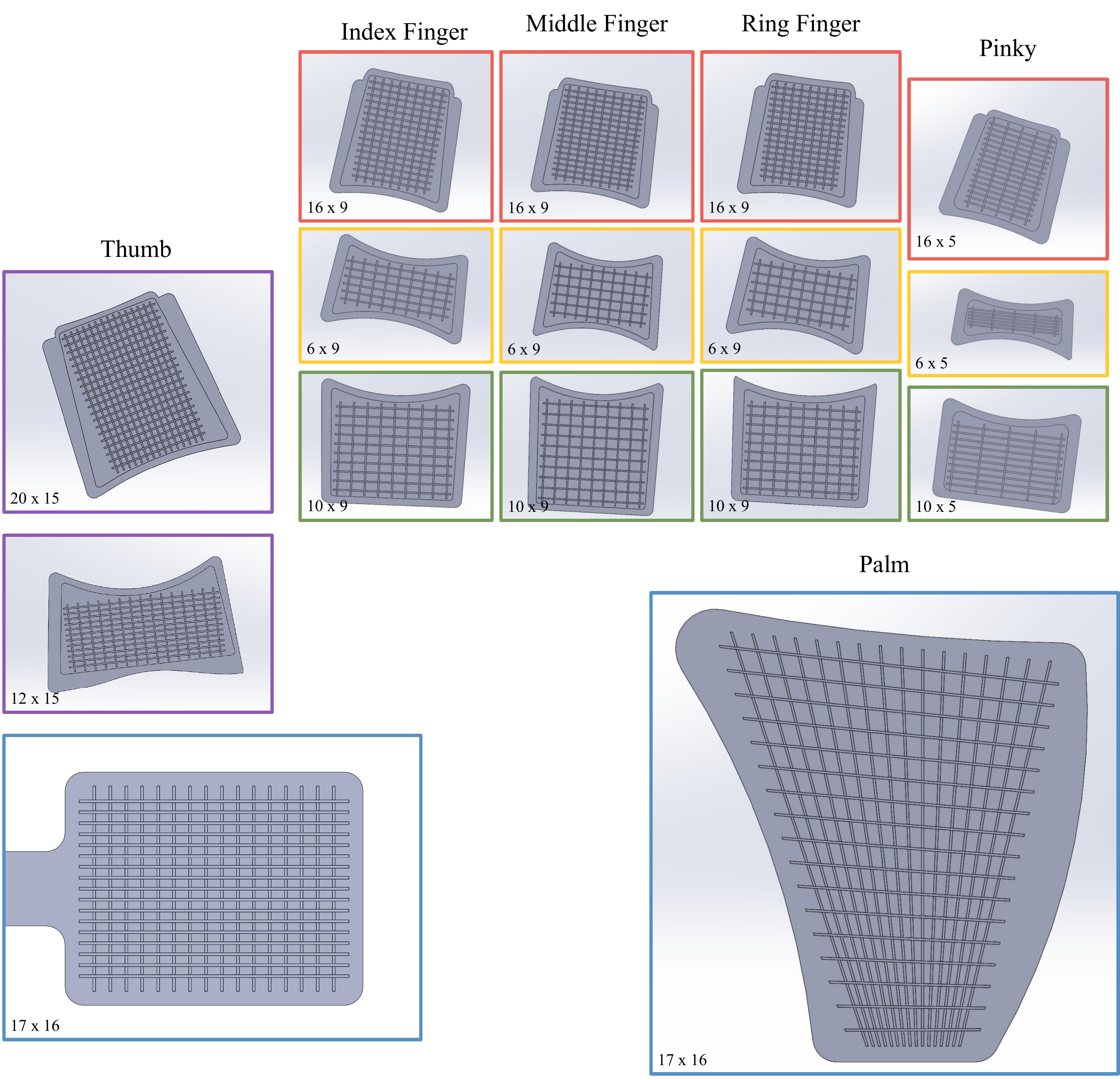}
\caption{\small
Taxel layout of the full-hand tactile skin. The $64 \times 32$ taxel array is distributed across the finger, thumb, and palm functional surfaces using three row-sharing groups. }
\label{fig:tactile_layout}
\end{figure}

As illustrated in Fig.~\ref{fig:tactile_layout}, the tactile skin of \nickname contains $64 \times 32 = 2048$ taxels distributed across the 16 rigid functional surfaces. To preserve a structured row--column readout, the layout is organized into three row-sharing groups. The four fingers share a $32 \times 32$ region. The thumb metacarpal surface and palm share a $17$-row region, with each surface occupying $17 \times 16$ taxels. The thumb distal and proximal surfaces share the remaining $15$-row region, with the columns split between the two surfaces. This organization keeps surfaces within the same group on common row lines while allocating different column spans according to surface size. The conductive traces use a width of $0.3\,\mathrm{mm}$.

\section{Encoder PCBs}

\begin{figure}[h]
\centering
\includegraphics[width=\linewidth]{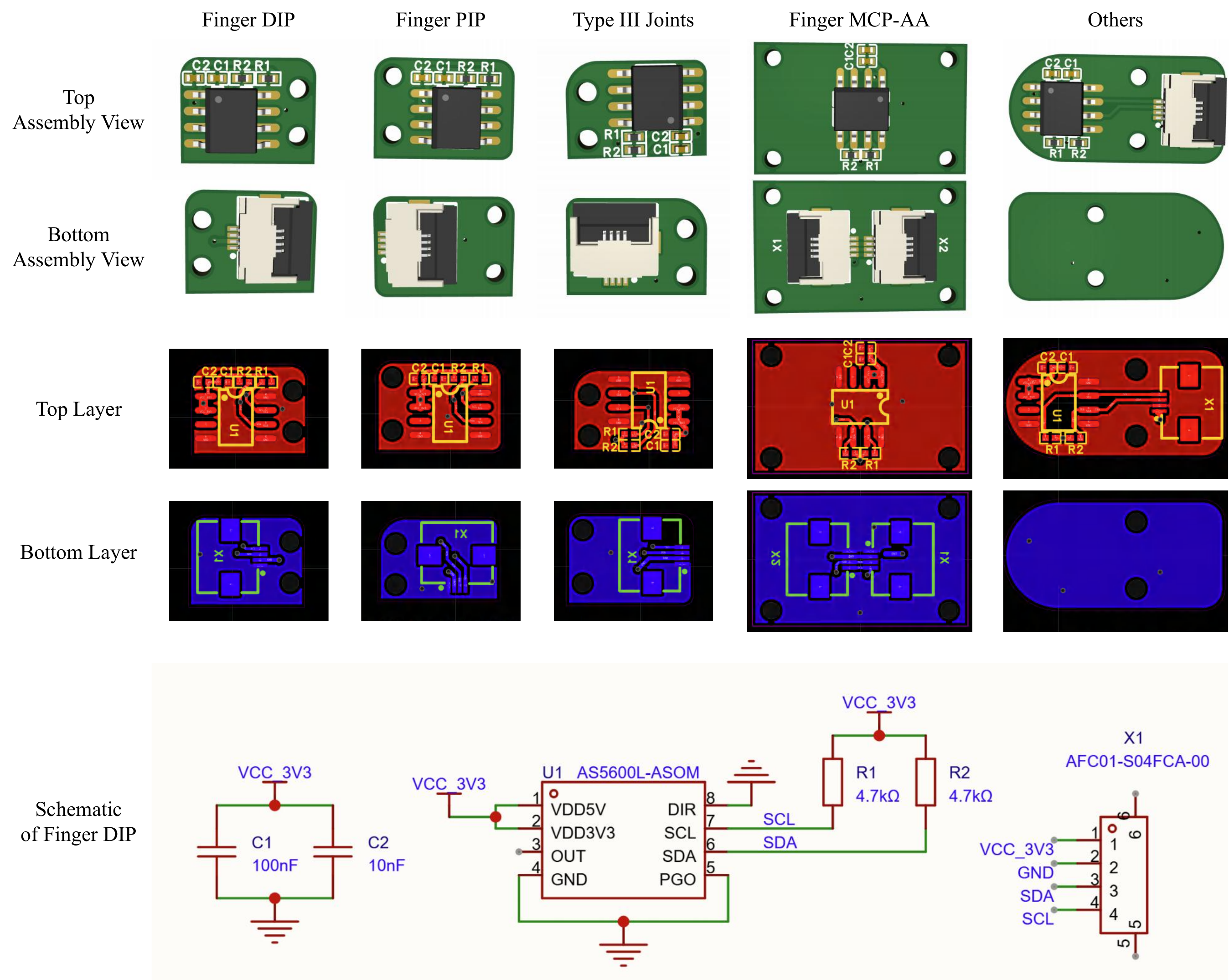}
\caption{\small Encoder PCB layouts of \nickname. Five PCB layouts are used for different joint locations and mounting constraints: finger DIP, finger PIP, Type III arc-slot joints, finger MCP-AA, and other one-side-mounted joints. The figure shows top and bottom assembly views, top and bottom copper layers, and a representative Finger DIP schematic.}
\label{fig:encoder_PCB}
\end{figure}

As shown in Fig.~\ref{fig:encoder_PCB}, \nickname uses five encoder-PCB layouts to accommodate different joint mechanisms and local mounting constraints. These include 4 PCBs for the finger DIP joints, 4 PCBs for the finger PIP joints, 5 PCBs for the Type III arc-slot joints, 4 PCBs for the finger MCP-AA joints, and 5 PCBs for the remaining joints, including the pinky MCP-FE, thumb IP, thumb MCP-FE, thumb MCP-AA, and thumb CMC-FE joints. Together, these layouts cover all 22 sensed DoFs.

All encoder PCBs use a two-layer design and communicate through a 4-pin $0.5\,\mathrm{mm}$-pitch FFC connector. The connector carries power, ground, SDA, and SCL signals for I2C communication. The representative Finger DIP schematic shows the common circuit shared across the encoder PCBs, including the AS5600L sensor, decoupling capacitors, and I2C pull-up resistors. This modular PCB design allows the same sensing principle to be reused across different joint mechanisms while preserving compact routing around the articulated glove.

\clearpage
\section{Readout Electronics}
\begin{figure}[h]
\centering
\includegraphics[width=\linewidth]{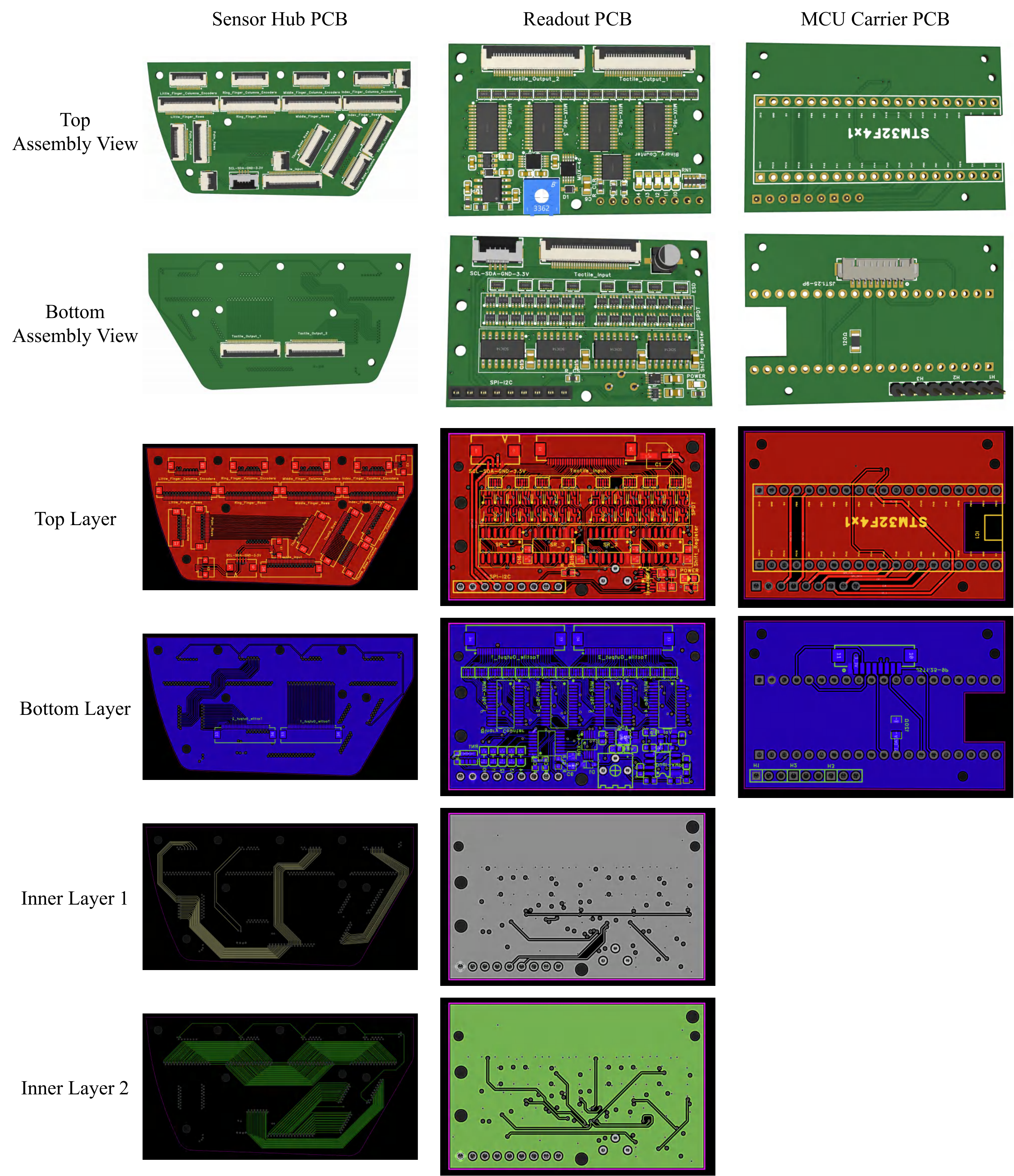}
\caption{\small Readout electronics of \nickname. The system consists of a
sensor hub, readout PCB, and MCU board for routing tactile and encoder signals
and acquiring synchronized data.}
\label{fig:pcbs}
\end{figure}

Fig.~\ref{fig:pcbs} shows the readout electronics of \nickname. The electronics stack consists of three PCBs: the sensor hub, the readout PCB, and the MCU board. The sensor hub collects tactile and encoder connections from the glove. The readout PCB scans the tactile array using a low-crosstalk readout circuit. The MCU board then acquires tactile measurements through SPI and encoder measurements through I2C for synchronized recording.

\end{document}